\pdfoutput=1
\pdfoutput=1

\documentclass[11pt]{article}

\usepackage{acl}

\usepackage{times}
\usepackage{latexsym}

\usepackage[T1]{fontenc}

\usepackage[utf8]{inputenc}

\usepackage{microtype}

\usepackage{inconsolata}

\usepackage{amsmath}
\usepackage{amssymb}
\usepackage{booktabs}
\usepackage{color}
\usepackage{enumitem}
\usepackage{graphicx}
\usepackage{multirow}
\usepackage{xspace}

\newcommand{\llama}{Llama~2\xspace}
\newcommand{\modelname}{KnowLA\xspace}
\newcommand{\conceptnet}{ConceptNet\xspace}
\newcommand{\wordnet}{WordNet\xspace}
\newcommand{\wikidata}{Wikidata\xspace}
\newcommand{\alpaca}{Alpaca2\xspace}
\newcommand{\commonsenseqa}{CommonsenseQA\xspace}
\newcommand{\siqa}{SIQA\xspace}
\newcommand{\triviaqa}{TriviaQA\xspace}
\newcommand{\bbh}{BBH\xspace}
\newcommand{\webquestionsp}{WebQuestionSP\xspace}
\newcommand{\truthfulqa}{TruthfulQA\xspace}
\newcommand{\kaping}{KAPING\xspace}
\newcommand{\contriever}{Contriever\xspace}

\title{KnowLA: Enhancing Parameter-efficient Finetuning with\\ Knowledgeable Adaptation}

\author{ 
    Xindi Luo$^\dagger$ \quad 
    Zequn Sun$^{\dagger,\,*}$ \quad
    Jing Zhao$^{\S,\,*}$ \quad 
    Zhe Zhao$^{\S,\,*}$ \quad
    Wei Hu$^{\dagger,\,\ddagger,\,}$\thanks{\,\,Corresponding authors} \\
    $^\dagger$ State Key Laboratory for Novel Software Technology, Nanjing University, China \\
    $^\ddagger$ National Institute of Healthcare Data Science, Nanjing University, China \\
    $^\S$ Tencent AI Lab, China \\
    \texttt{xdluo.nju@gmail.com,\,sunzq@nju.edu.cn} \\ 
    \texttt{\{janinezhao,nlpzhezhao\}@tencent.com,\,whu@nju.edu.cn}}

\begin{document}
\maketitle
\begin{abstract}
Parameter-efficient finetuning (PEFT) is a key technique for adapting large language models (LLMs) to downstream tasks.
In this paper, we study leveraging knowledge graph embeddings to improve the effectiveness of PEFT. 
We propose a knowledgeable adaptation method called KnowLA. 
It inserts an adaptation layer into an LLM to integrate the embeddings of entities appearing in the input text. 
The adaptation layer is trained in combination with LoRA on instruction data.
Experiments on six benchmarks with two popular LLMs and three knowledge graphs demonstrate the effectiveness and robustness of KnowLA. 
We show that \modelname can help activate the relevant parameterized knowledge in an LLM to answer a question without changing its parameters or input prompts.
\end{abstract}

\section{Introduction}

In the era of large language models (LLMs) with billions and possibly trillions of parameters~\cite{GLM,GPT4,llama1}, parameter-efficient finetuning (PEFT) stands out as a crucial technique enabling the necessary adaptation of LLMs to downstream tasks.
It freezes most or even all parameters of LLMs and only finetunes a small number of parameters using limited instruction data.
LoRA~\cite{LORA} is a widely-used PEFT method that trains small low-rank adapters to approximate the large layers in LLMs.
Follow-up work improves the efficiency of LoRA by using quantized weights \cite{qlora}.
In this work, we seek to improve the effectiveness of LoRA while preserving comparable efficiency.

Inspired by knowledge-injected pre-trained language models (PLMs), e.g., ERNIE~\cite{ERNIE}, 
we explore knowledge graphs (KGs) to enhance the PEFT of LLMs with LoRA. 
A KG is a large-scale structured knowledge base containing a massive amount of trustworthy knowledge. 
The typical way of injecting KGs into PLMs in the past few years is incorporating pre-trained entity embeddings at the input layer of a PLM and finetuning the full model on NLP tasks ~\cite{LIBERT, KnowBERT, KT-NET, ERNIE, SenseBERT, KG-BART, KELM, EDR}.
Knowledge injection has improved many PLMs, e.g., BERT~\cite{bert} and RoBERTa~\cite{Roberta}. 
However, previous knowledge injection methods require fully tuning PLMs, which is inapplicable to LLMs.
Furthermore, these methods are founded on the encoder-based architecture of PLMs, and their effectiveness for recent decoder-based LLMs remains unknown. 
The following questions thereby arise: \textit{Can knowledge injection still enhance the PEFT of LLMs? Also, how can knowledge injection be used to enhance PEFT?} 

To answer these questions, in this paper, we propose a knowledgeable adaptation method for PEFT, particularly for LoRA, called \modelname.
It inserts an adaptation layer into a pre-trained LLM.
The layer integrates external KG embeddings of entities appearing in the input text of the LLM.
Entity embeddings and parameters of the LLM are frozen in PEFT.
The proposed adaptation layer is trained combined with LoRA on instruction data.
The parameters in our adaptation layer are significantly fewer than those in the LLM and even fewer than those in LoRA.
Thus, our \modelname is also a parameter-efficient method without changing the original parameters of the LLM. 

We evaluate \modelname on six datasets, including commonsense reasoning on \commonsenseqa~\cite{csqa}, social interaction reasoning on \siqa~\cite{siqa} and BIG-Bench Hard \cite{bbh}, single-hop reasoning of KBQA on \webquestionsp~\cite{wqsp}, and close-book QA on \triviaqa \cite{triviaqa} and \truthfulqa \cite{truthfulqa}.
Experimental results show that \modelname can enhance the effectiveness of LoRA at the expense of a limited number of additional parameters. 
Even when compared to Alpaca2 \cite{alpaca}, which has a larger LoRA with a similar number of parameters, \modelname with a smaller LoRA achieves better results.

We assess the robustness of \modelname with two popular foundation models (i.e., LLaMA 1 \cite{llama1} and Llama 2 \cite{Llama2}), different instruction data (i.e., instruction-following demonstrations in Alpaca2 and Vicuna2 \cite{vicuna2023}), various KGs (i.e., WordNet \cite{WordNet}, ConceptNet~\cite{ConceptNet}, and Wikidata~\cite{wikidata}), and typical embedding learning models (i.e., RESCAL \cite{rescal}, TransE~\cite{TransE}, and RotatE \cite{rotate}), combined with two PEFT methods (i.e., LoRA~\cite{LORA} and AdaLoRA~\cite{adalora}).
Experiments show that \modelname can offer stable improvements.

To understand how \modelname changes the output of an LLM, we analyze the results from two perspectives, which show several interesting findings: 
(i) \modelname with LoRA can align the space of the LLM with the space of KG embeddings, and 
(ii) \modelname can activate the parameterized potential knowledge that originally exists in the LLM, even though the used KG does not contain such knowledge.
According to our findings, in some cases, the LLM outputs incorrect answers not because it does not know the answers, but because its relevant knowledge is not activated by the input prompts.
\modelname can help activate its relevant knowledge without changing its parameters or input prompts.

\section{Related Work}

\subsection{Knowledge Injection}
There are three typical knowledge injection methods for PLMs.
The first method involves KG embeddings at the input layer of PLMs for joint learning~\cite{ERNIE,KELM,Kepler}. 
Existing works incorporate entity embeddings for classification tasks, and their knowledge injection modules are independent of PLMs. 
This poses challenges to aligning the semantic spaces of entity embeddings and PLMs.
These knowledge injection methods also necessitate updating the entire model of PLMs.
The second method converts relevant triples in KGs into natural language sentences used for pre-training PLMs~\cite{K-BERT,CoLAKE,ERNIE-Baidu}. 
The third method introduces adapters into PLMs to enable them to learn KGs \cite{K-Adapter}.
Our \modelname relates to the first type of methods. 
It is also a variant of the third method. 
However, previous methods are built on PLMs while our method is the first attempt to LLMs. 
\modelname does not update the parameters of LLMs.
It employs a knowledge adapter during PEFT to enhance the LLM's capabilities.
The injected entity knowledge can also be deeply integrated with the LLM's knowledge in subsequent decoding steps.

Apart from the above work injecting knowledge inside the model, there are also methods retrieving and augmenting relevant knowledge on the input side of the model \cite{self-talk, Contriever, GKP, KAPING}. 
For example, given an input, \contriever \cite{Contriever} extracts relevant passages from Wikipedia. 
GKP \cite{GKP} generates relevant prompt text using a sophisticated LLM.
\kaping \cite{KAPING} retrieves relevant triples in KGs. 

\begin{figure}
\centering
\includegraphics[width=\columnwidth]{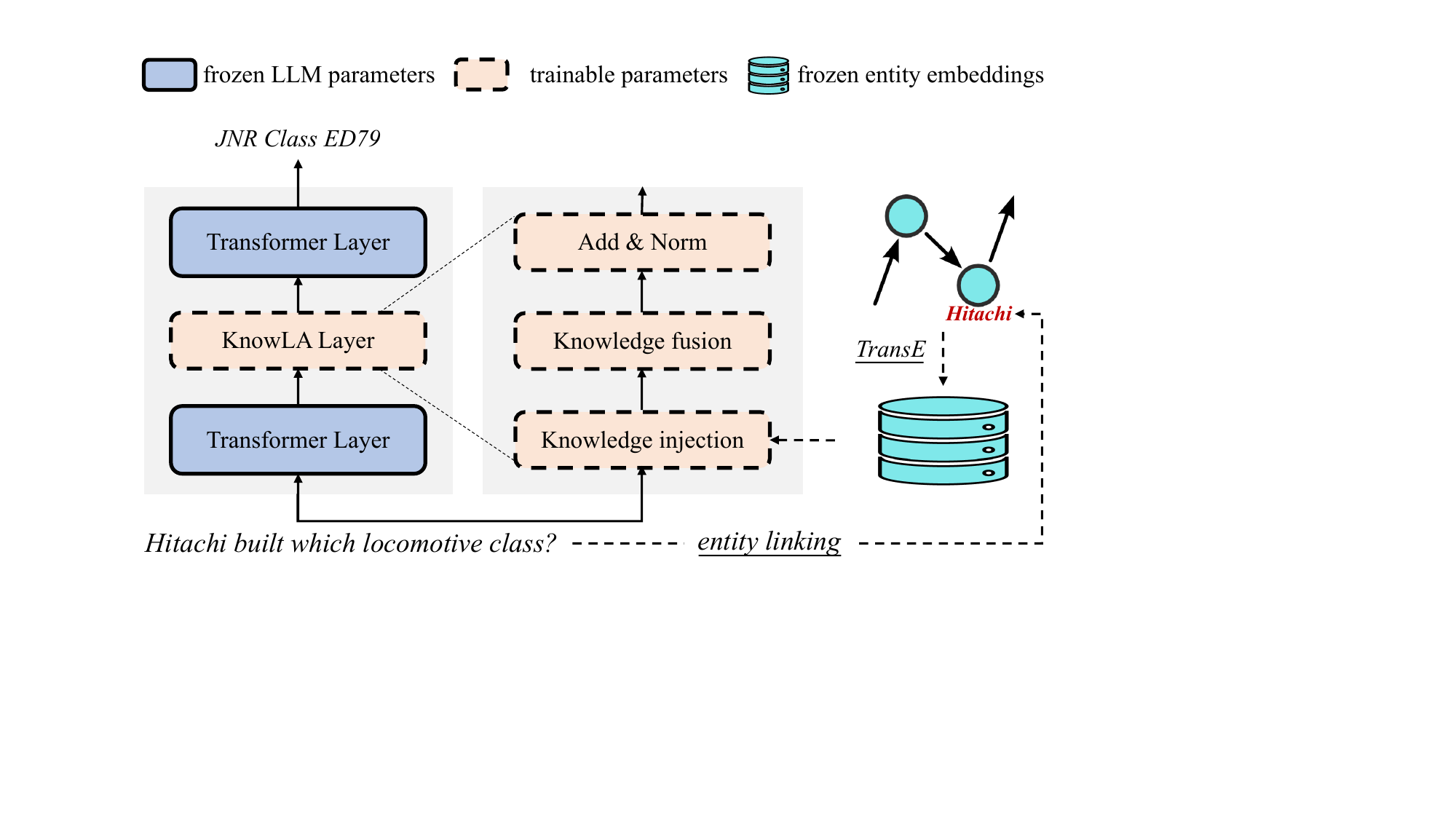}
\caption{Illustration of knowledgeable adaptation. 
The \modelname layer is inserted between two decoder layers of an LLM. 
It consists of knowledge injection and fusion.}
\label{fig:kia}
\end{figure}

\subsection{Parameter-efficient Finetuning}
PEFT methods aim to optimize LLMs while minimizing the computational resources and data required. 
Adapter Tuning \cite{adapter} is a lightweight alternative that inserts a small neural module called adapter in each layer of a PLM while keeping the majority of the pre-trained parameters frozen. 
Inspired by the prompt engineering methods, Prefix Tuning \cite{Prefix-Tuning} sets trainable prefix tokens in the input or hidden layers, and only these soft prompts are trained. 
LoRA~\cite{LORA} is a low-rank adaptive method that allows training dense layers indirectly by optimizing low-rank factorized matrices that capture changes in dense layers during the adaptation process while keeping the pre-trained weights unchanged. 
QLoRA~\cite{qlora} improves LoRA by using NF4 quantization and double quantization techniques.
Adalora \cite{adalora} is an improvement on LoRA, addressing the limitation of the fixed incremental matrix rank in LoRA.
Adalora introduces a method that dynamically allocates ranks for downstream tasks, yielding promising results.
Our \modelname follows the mainstream research of LLMs and achieves PEFT with fewer parameters combined with LoRA. 
During the finetuning process, the parameters of LLMs and entity embeddings are fixed, allowing only gradient backpropagation through the parameters of adapters. 
This enables the use of external knowledge to unleash the potential of LLMs. 

\section{\modelname}

Considering that the hidden states in Transformer layers encapsulate the parameterized knowledge of an LLM \cite{pmet},
we propose fusing entity embeddings in a KG with the hidden states of an LLM during PEFT. 
\modelname inserts an adaptation layer into an LLM, as shown in Figure~\ref{fig:kia}. 

Given a KG, we adopt a representation learning model, e.g., TransE \cite{TransE}, to train its entity embeddings.
The pre-trained embedding of entity $e$ is denoted by $\mathbf{e}$.
For an input question $Q = {\{t_i\}}^{n}_{i=1}$ to an LLM, each token $t_i$ may be linked to a set of entities $E(t_i)$ in the KG.
Our key idea is to enhance PEFT by injecting the embedding $\mathbf{e}_i$ for each $e_i \in E(t_i)$ into the representation in the LLM. 
This method can be divided into three modules: 
(i) \textit{Entity linking}, which links the tokens in a question to entities in the KG.
(ii) \textit{Knowledge mapping and injection}, which maps the KG embedding space to the LLM's representation space and infuses the entity embeddings corresponding to a specific token in the question. 
(iii) \textit{Knowledge fusion}, which integrates each token representation with its entity embedding. 
Given the powerful abilities, popularity, and open-source nature of the LLaMA family~\cite{llama1,Llama2}, we consider it the foundation to build our \modelname.

\subsection{Entity Linking} 
Given an input text, we return its synsets as candidate entities in a KG. 
We use the text-rank algorithm to recognize important tokens and link the recognized tokens to the KG by string matching. 
We also collect a set of synonyms for each related entity. 
Based on the byte pair encoding (BPE) algorithm \cite{BPE}, each token is divided into multiple subwords sharing the same entity candidate. 
After this step, we obtain relevant entities in the KG for the important tokens in the text. 
Each entity is associated with a pre-trained embedding.

\subsection{LLM Encoding}
Given an LLM, e.g., \llama, it first encodes the input text to get embeddings for prompts and questions. 
Specifically, for a prompt $p$, the LLM first converts it into $Q$ = $([s],p,[/s])$. 
The decoder of the LLM tokenizes $Q$ with the BPE algorithm. 
After tokenization, $Q$ turns into ${\{\mathbf{h}_i \}}^{m}_{i=1} \in \mathbb{R}^{d_1}$, which is taken as input to the LLM.

\subsection{Knowledge Mapping and Injection}
The text representation of the $l$-th decoder layer in the LLM is denoted by $\mathbf{h}^l$. 
In the knowledge mapping module, to align with the pre-norm mode adopted by the decoder and mitigate the issues of gradient vanishing or exploding, we apply RMSNorm~\cite{RMSNorm} to the input $\mathbf{h}^l$ received by the decoder. 
We also map the semantic space of entity embeddings to the semantic space of the LLM for transformation, aiming to improve knowledge injection and fusion.

The BPE encoding method employed by many LLMs would let each token have multiple sub-tokens after encoding. 
Let ${\{\mathbf{h}^l_i\}}^{k}_{i=1}$ denote the sub-token embeddings, where $k$ is the number. 
To better calculate the relevance between different entities and the given word, we unify the representations of the $k$ sub-tokens as $\mathbf{u}_i$ using mean pooling:
\begin{equation}
\mathbf{u}_i = \mathrm{AvgPooling}(\mathbf{h}^l_1, \dots, \mathbf{h}^l_k) .
\end{equation}

As LLMs are employed for handling complex natural language tasks, it is essential to have input dimensions sufficiently large to accommodate the intricacies. 
To enhance the expressive ability of entity representation $\mathbf{e}_i$ and align with the semantic space of the LLM, we expand its dimension to enrich the representation of $\mathbf{e}_i$:
\begin{equation}
\mathbf{e}_i = \mathbf{W}_{d}\big(\mathrm{SwiGLU}(  \mathbf{W}_{u}\,\mathbf{e}_i + \mathbf{b}_u)\big),
\end{equation}
where $\mathbf{W}_{d}\in \mathbb{R}^{d_1 \times d_3}$, $\mathbf{W}_{u} \in \mathbb{R}^{d_3 \times d_2}$, and $\mathbf{b}_{u} \in \mathbb{R}^{d_3}$ are trainable weights. 
SwiGLU~\cite{swiglu} is an activation function. 

\subsection{Knowledge Fusion}   
To mitigate the risk of the LLM encountering unfamiliar entities during finetuning in downstream tasks, as well as to ensure the extracted entities are relevant to the input tokens, we follow \cite{KT-NET} and introduce a knowledge sentinel $\overline{\mathbf{e}}$. 
First, we calculate the similarities of each token with its relevant entities and the knowledge sentinel:
\begin{align}
\alpha_{ij} &= \frac{{\exp}(\mathbf{e}_j \cdot \mathbf{u}_i)}{\sum_{j}{\exp}(\mathbf{e}_j \cdot \mathbf{u}_i) + {\exp}(\overline{\mathbf{e}} \cdot \mathbf{u}_i)},\\
\beta_i &= \frac{{\exp}(\overline{\mathbf{e}} \cdot \mathbf{u}_i)}{\sum_{j}{\exp}(\mathbf{e}_j \cdot \mathbf{u}_i) + {\exp}(\overline{\mathbf{e}} \cdot \mathbf{u}_i)},
\end{align}
where $\alpha_{i j}$ represents the relevance between the $i$-th token and the $j$-th entity. 
$\beta_i$ represents the relevance between the $i$-th token and the knowledge sentinel. 
Here, we constrain that $\sum_j \alpha_{i j}+\beta_i=1$. 
Then, we fuse $\mathbf{u}_i$ with its relevant entities:
\begin{align}
\overline{\mathbf{u}}_i&=\sum_j \alpha_{i j}\,\mathbf{e}_j+\beta_i\,\overline{\mathbf{e}},\\
\overline{\mathbf{h}}_i&=\theta\,\mathrm{SwiGLU}\big(\mathbf{W}_m [\overline{\mathbf{u}}_i;\mathbf{u}_i]+ \mathbf{b}_m\big) + \mathbf{h}_i,
\end{align}
where $\theta$ serves as a trainable balancing factor to equalize the impact of KG and text. $\mathbf{W}_{m} \in \mathbb{R}^{2d_1 \times d_1}$ and $\mathbf{b}_{m} \in \mathbb{R}^{d_1}$ are trainable weights. 
During knowledge fusion, all the $k$ sub-token embeddings ${\{\mathbf{h}_i\}}^{k}_{i=1}$ share the same $\overline{\mathbf{u}}_i$. 
$\overline{\mathbf{h}}_i$ denotes the final representation of knowledge injection and serves as the output of the current adapter, which is passed as input to the next layer of the decoder.

Similar to other parameter-efficient modules like LoRA \cite{LORA}, \modelname achieves the alignment between KG knowledge and textual semantics by freezing the LLM during finetuning. 
It can also be used in conjunction with LoRA to achieve efficient learning of the LLM with a limited number of parameters. 
The effectiveness of this module is shortly assessed in the experiments.

\section{Experiments}

\subsection{Baselines and Implementation}
We consider the following LLMs with 7B parameters as foundation models in our main experiments:
\begin{itemize}[itemsep=1pt,topsep=2pt]
    \item \textbf{\llama} is a collection of open-source LLMs trained on public datasets with trillions of tokens. 
    We use the \llama-7B model.

    \item \textbf{Alpaca2} \cite{alpaca} is a \llama variant finetuned with 52,000 instruction-following demonstrations using LoRA.
\end{itemize}

Given that there are currently no knowledge injection methods for PEFT, we choose retrieval augmented generation (RAG) methods as baselines:
\begin{itemize}[itemsep=1pt,topsep=2pt]
    \item \textbf{\contriever} \cite{Contriever} is pre-trained using English Wikipedia. 
    We use it to retrieve triples from KGs and passages from Wikipedia to augment the input of the LLM.
    
    \item \textbf{\kaping} \cite{KAPING} retrieves relevant triples from KGs to improve the KBQA task. 
    We use \kaping to enhance LLMs on knowledge-relevant tasks.
\end{itemize}

In our main experiments, we use the official hyperparameters and instruction data of \alpaca to finetune Llama 2-7B with LoRA and \modelname.
Our layer is inserted after the 32nd layer of \llama. 
We also consider LLaMA~1 and the instruction data of Vicuna2 \cite{vicuna2023} in Sect. (\ref{sect:other}).

During the training process, we set the batch size to 128 and the learning rate to 3e-4, and use the AdamW optimizer to train 3 epochs. 
We keep the hyperparameters the same for different models to ensure the fairness of the experiment. 
We also keep the input prompts the same in the experiments. 
To study the impact of the number of trainable parameters, we train two LoRA models with different ranks: $r=16$ and $32$. 
They both perform better than ranks $r=4,8$ on most datasets.
All models are finetuned on A800 GPUs.
The code is publicly available at our GitHub repository.\footnote{\url{https://github.com/nju-websoft/KnowLA}}

\subsection{Datasets and Settings}
We consider three types of tasks: multi-choice QA, closed-book QA, and truthful QA. 
We pick \commonsenseqa \cite{csqa} and SIQA \cite{siqa} as the multiple-choice QA datasets, and choose 15 challenging multi-choice tasks from BIG-Bench Hard (BBH) \cite{bbh}. 
We use \webquestionsp \cite{wqsp} and TriviaQA \cite{triviaqa} for closed-book QA evaluation. 
We also use TruthfulQA \cite{truthfulqa} to evaluate whether \modelname is truthful in generating answers to questions. 
Appendix~\ref{app:datasets} complements more details.
To assess the direct improvement of our \modelname to enhance PEFT, we employ zero-shot settings for all tasks. 

\begin{table*}[!t]
\centering
\resizebox{0.8\linewidth}{!}{
\begin{tabular}{l|c|cccccc}
\toprule
\multirow{2}{*}{Methods} & \multirow{2}{*}{\#Parameters} & \multicolumn{2}{c}{CommonsenseQA} & \multicolumn{2}{c}{SIQA} & \multicolumn{2}{c}{BIG-Bench Hard} \\
\cmidrule(lr){3-4}\cmidrule(lr){5-6}\cmidrule(lr){7-8}
& & Accuracy & Score & Accuracy & Score & Accuracy & Score \\ 
\midrule
\llama (7B) & 7B & 45.37	&36.40	&46.42	&40.58	&26.95	&24.87\\
\alpaca ($r=16$) & +\,0.24\% & 56.92	&46.55	&52.61	&46.18	&28.93	&\textbf{25.42}\\
\alpaca ($r=32$) & +\,0.50\% & 57.90	&46.81	&53.17	&46.21	&28.79	&25.36\\
\midrule
\contriever (\wordnet)& \multirow{2}{*}{+\,0.50\%} & 57.15	&46.09	&52.58	&46.13	&-	&-\\
\contriever (\conceptnet)&  & 57.06	&45.30	&52.51	&45.51	&-	&-\\
\kaping (\wordnet)& \multirow{2}{*}{+\,0.50\%} & 57.21	&45.91	&52.51	&45.89	&-	&-\\
\kaping (\conceptnet)&  & 57.58	&45.64	&52.66	&46.15	&-	&-\\
\midrule
\modelname (Random) & \multirow{4}{*}{+\,0.55\%} & 57.49 &47.82	&52.61	&46.56	&29.26	&25.34\\
\modelname (\wordnet) &  & 58.07 &\textbf{48.35} &\textbf{53.22} &46.76	&30.00 &25.39\\
\modelname (\conceptnet) &  &\textbf{58.39} &48.19	&\textbf{53.22} &\textbf{46.81} &\textbf{30.19} &25.29 \\
\modelname (\wikidata) &  &57.90 &47.39 &53.21 &46.64 &29.39 &\textbf{25.42}\\
\bottomrule
\end{tabular}}
\caption{Multi-choice QA results on \commonsenseqa, \siqa, and \bbh. For \modelname, the rank of LoRA is $r=16$. The percentage of trainable parameters are similar in Tables~\ref{tab:triviaqa} and \ref{tab:truthfulqa}.}
\label{tab:commonsenseqa}
\end{table*}

\subsection{KGs and Configurations}
We select \wordnet \cite{WordNet}, \conceptnet \cite{ConceptNet}, and \wikidata~\cite{wikidata} as the KGs in our method.
See Appendix~\ref{app:datasets} for more descriptions.

For RAG methods, we consider the overlap between questions and knowledge sources. 
For multi-choice QA, we use \conceptnet and \wordnet. 
For \triviaqa, we use \wikidata and Wikipedia.

For KG embeddings, we follow~\cite{ERNIE} and pre-train entity embeddings with TransE \cite{TransE} as the external knowledge. 
The maximum number of relevant entities selected for each textual token in a question is set to 5. 
Furthermore, we evaluate the side effects and additional latency of \modelname. See Appendix~\ref{app:side_effect} and Appendix \ref{app:efficiency} for more details.

\subsection{Experiments on Multi-choice QA} \label{csqa}
To evaluate the effectiveness and robustness of \modelname, we compare it to Llama 2 and Alpaca2 ($r=16,32$) on multi-choice QA. 
In addition to accuracy, we follow~\cite{self-talk} and compute scores using cross entropy, which indicate the confidence of a model for correct answers.
We use three KGs: \wordnet, \conceptnet, and \wikidata.
We also consider randomly initialized vectors as a baseline of KG embeddings.

Table \ref{tab:commonsenseqa} presents the results. 
Our \modelname variants show the best performance across the three datasets. 
Furthermore, Alpaca2 ($r=32$) outperforms Alpaca2 ($r=16$), because more trainable parameters usually lead to better performance. 

\kaping generally performs better than \contriever on \commonsenseqa. 
This indicates that the RAG methods rely on the quality of prompts retrieved from the knowledge sources. 
Both \kaping and \contriever are inferior to Alpaca2 ($r=32$) on \commonsenseqa and \siqa, as invalid prompts may cause damage to the performance.

\modelname is different from RAG methods. 
RAG methods retrieve text information to augment the input of LLMs, while \modelname uses KG embeddings to improve the effectiveness of PEFT. 
\modelname works in the finetuning phase of LLMs and does not change the input of LLMs. 
Our method with LoRA ($r=16$) achieves better performance than all baselines, indicating that it can effectively work with PEFT to inject knowledge.
Specifically, when combined with \conceptnet, it achieves an accuracy increase from $56.92\%$ to $58.39\%$ on \commonsenseqa, from $52.61\%$ to $53.22\%$ on \siqa, and from $28.93\%$ to $30.19\%$ on BBH. 
Since \conceptnet stores rich conceptual knowledge and more relation types compared to \wordnet, its entity embeddings can better enhance \llama's reasoning ability. 
Furthermore, \conceptnet recognizes more relevant entities in the question than \wikidata.
This suggests that extensive entity coverage in \modelname brings a significant performance increase.

Additionally, the performance of \modelname (random) is inferior to that of \modelname with KGs, highlighting the greater utility of entity knowledge for LLMs.
Based on the scores of each model on the correct answers, it can be seen that after incorporating \modelname, all models assign higher confidence to the correct answers. 
Therefore, \modelname can offer a certain degree of improvement for LLMs in commonsense reasoning.

\begin{figure*}[!t]
\centering
\includegraphics[width=0.999\linewidth]{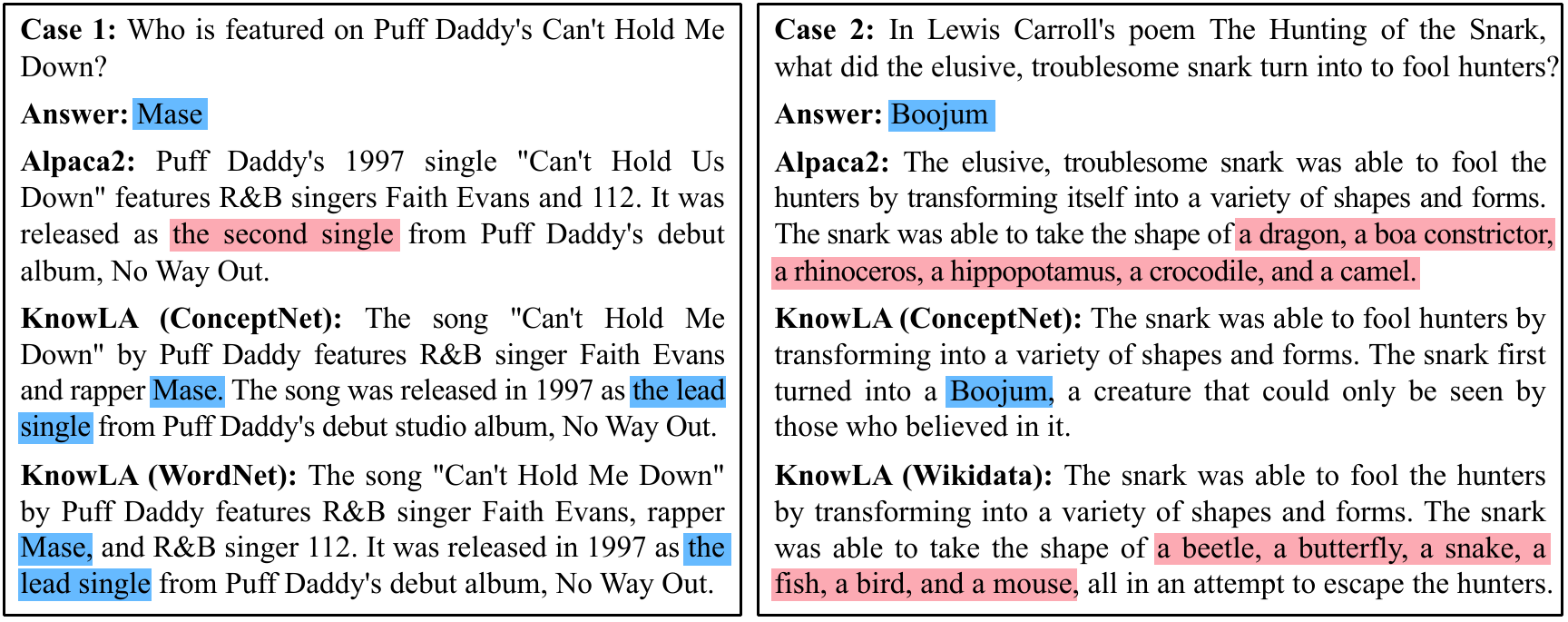}
\caption{Examples of \alpaca and \modelname for \triviaqa.}
\label{fig:case_study}
\end{figure*}

\begin{table}
\centering
\resizebox{.94\columnwidth}{!}{
\begin{tabular}{l|cc}
\toprule
Methods & \webquestionsp & \triviaqa \\ 
\midrule
\alpaca ($r=16$) & 67.55 & 68.70 \\
\alpaca ($r=32$) & 67.43 & 67.97\\
\midrule
\contriever (Wikipedia) & - & 68.71 \\
\kaping (\wikidata) & 67.11 & 66.05 \\
\midrule
\modelname (Random) & 67.68 & 69.34\\
\modelname (\wordnet) & 67.43 & 69.27\\
\modelname (\conceptnet) & \textbf{68.12} & \textbf{69.40}\\
\modelname (\wikidata) & 67.49 & 68.92\\
\bottomrule
\end{tabular}}
\caption{Closed-book QA results on \webquestionsp and \triviaqa. For \modelname, $r=16$.}
\label{tab:triviaqa}
\end{table}

\subsection{Experiments on Closed-book QA} \label{cbqa}
We evaluate \modelname using \webquestionsp and \triviaqa. 
Following the answer matching strategy in \cite{KBQA_eval}, we use the subtree labels provided by the constituent tree to extract all noun phrases from the textual answers, calculate their similarities, and determine the correctness of answers exceeding a certain threshold (e.g., 0.5).

The accuracy results are shown in Table~\ref{tab:triviaqa}. 
We find that Alpaca2 ($r=16$) obtains better performance than Alpaca2 ($r=32$). 
The reason may be that more parameters in LoRA are prone to overfitting in the closed-book QA tasks.
Moreover, \contriever (Wikipedia) only slightly exceeds Alpaca2 ($r=16$) and performs better than \kaping. 
This is because \kaping cannot guarantee the correctness of the extracted triples. 

According to the results, \modelname combined with \wordnet improves the results from $68.70\%$ to $69.27\%$ on \triviaqa, while combined with \conceptnet, the performance is further enhanced to $69.40\%$. 
This indicates that the parameterized entity embeddings can enrich the textual representations. 
The experimental results demonstrate that the knowledge-enhanced textual representations after finetuning with LoRA can help mitigate the hallucination problem of \llama to some extent.

On \webquestionsp, \modelname (\wordnet) and \modelname (\wikidata) produce similar results. 
Also, the two Alpaca2 models with different ranks perform similarly. 
This suggests that the reasoning ability of Alpaca2 is good on this task, and the performance does not change significantly after knowledge enhancement with \modelname. 
We attribute this bottleneck to the model size and the training data of \llama and Alpaca2. 

\subsection{Experiments on TruthfulQA} \label{truthfulqa}
We use \truthfulqa to measure whether \modelname is truthful in generating answers to questions. 
Here, we evaluate the content generated by the models based on the best answer provided by \truthfulqa, using the commonly used metrics BLEU, Rouge-1, Rouge-2, and Rouge-L. 
Table~\ref{tab:truthfulqa} shows the results. 

\begin{table}
\centering
\resizebox{\columnwidth}{!}{\Large
\begin{tabular}{l|cccc}
\toprule
Methods & BLEU & Rouge-1 & Rouge-2 & Rouge-L \\ 
\midrule
\alpaca($r=16$) &0.1657 &0.4094 &0.2831 &0.3892\\
\alpaca($r=32$) &0.1637 &0.4048 &0.2802 &0.3851\\
\midrule
\modelname(Random) & 0.1677	& 0.4110 & 0.2850	&0.3897\\
\modelname(\wordnet) & 0.1714 &0.4143 &0.2874 &0.3927\\
\modelname(\conceptnet) & \textbf{0.1747} & \textbf{0.4190} & \textbf{0.2922} & \textbf{0.3975}\\
\modelname(\wikidata) & 0.1703 &0.4135 &0.2895 &0.3931\\
\bottomrule
\end{tabular}}
\caption{Results on \truthfulqa. For \modelname, $r=16$.}
\label{tab:truthfulqa}
\end{table}

Alpaca2 ($r=32$) still underperforms Alpaca2 ($r=16$). 
This further substantiates our conclusion that larger parameters do not necessarily guarantee the accuracy and reliability of the model's output. 
\modelname (\conceptnet) performs best among these models, which indicates that the integration of our \modelname with LoRA can mitigate the hallucination problem of \llama to some extent and generate content of better quality.

Besides, we observe that \modelname (\conceptnet) outperforms \modelname (\wordnet) in all evaluation tasks, and \modelname (\wordnet), in turn, surpasses \modelname (\wikidata). 
This further indicates that the commonsense knowledge within \conceptnet is more suitable for both LoRA and \llama.

\subsection{Case Study}
Figure \ref{fig:case_study} presents some improved results of Alpaca2 by incorporating \wordnet, \conceptnet, and \wikidata in \modelname.
In Case 1, we discover that after integrating \conceptnet and \wordnet with \modelname, the response precisely describes the correct answers. 
The contents generated by \modelname (\conceptnet) and \modelname (\wordnet) are very similar. 
The content generated by Alpaca2 not only misses significant answers but also misinterprets the song ``Can't Hold Me Down'' in the question. 
Therefore, we believe that \modelname helps the model better understand questions. 

By examining the answers of the three models in Case 2, it can be observed that \alpaca does not provide an accurate and relevant response, which is similar to the content generated by \modelname (\wikidata). 
They both generate deceptive answers. 
However, after incorporating \conceptnet, \modelname accurately provides the correct answer in the response. 
According to Table \ref{tab:triviaqa}, we believe that the enhancement is not accidental. 
Moreover, by examining the token-to-entity linking results, we find that \textit{the answer entity ``Boojum'' does not exist in \conceptnet.} 
Therefore, we conclude that \modelname can stimulate the underlying reasoning abilities of LLMs by working with LoRA.

\subsection{Why Knowledgeable Adaptation Works?} \label{work}
We delve into why \modelname collaborates effectively with LoRA, focusing on space alignment of KGs and LLMs, and knowledge recall in LLMs.

\begin{figure}[!t]
\centering
\includegraphics[width=\linewidth]{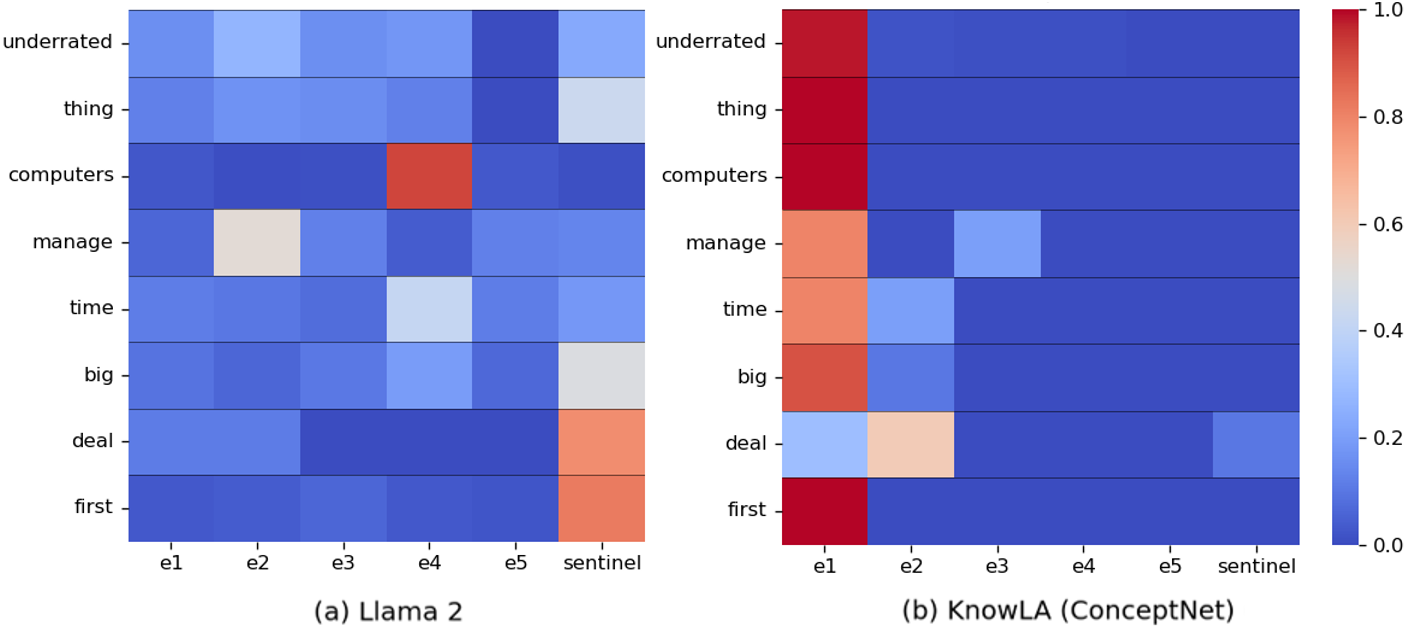}
\caption{The similarity heatmap between the output representations of text tokens and their corresponding entity embeddings. 
The x-axis denotes the top-5 similar entities with tokens on the y-axis.
(a) The left heatmap presents the similarity of Llama 2 without finetuning, and (b) the right heatmap presents the similarity after finetuning with our \modelname (ConceptNet).}
\label{fig:case_study_alignment}
\end{figure}

\paragraph{Perspective of Space Alignment.} 
Our \modelname incorporates pre-trained KG embeddings into a pre-trained LLM for instruction tuning with LoRA. 
We hereby investigate whether the two heterogeneous representation spaces of the KG and the LLM are aligned, to understand how \modelname works.
The results are illustrated in Figure \ref{fig:case_study_alignment}, where the last column represents the ``sentinel'' entity. 
We first acquire the representations of the input tokens in a specific layer, e.g., the 32nd layer.
Then, we retrieve the top five similar entity embeddings in the KG for each token. 
Next, to establish the relevance of each token and its corresponding entities, we calculate the attention weights between them. A larger weight suggests a stronger semantic correlation between the token and the mapped entity.

In the case of Llama 2 (depicted in the left part of Figure \ref{fig:case_study_alignment}), the similarities between entity embeddings and token representations appear to be random, lacking any discernible patterns.
However, after applying \modelname, 
the results show improved accuracy specifically for the most relevant entities (i.e., $e_1$ on the x-axis). 
For token ``underrated'', the relevant entities in ConceptNet are ``underrated'', ``underrate'', etc. 
After finetuning, the token ``underrated'' exhibits the highest correlation with the entity ``underrated''. 
This observation indicates that \modelname can effectively align the KG and the LLM through instruction tuning with LoRA. 

\paragraph{Perspective of Knowledge Recall.}
We study the role of \modelname in activating an LLM's knowledge. 
According to \cite{pmet, ffn1, ffn2}, the feed-forward network (FFN) layers, which constitute two-thirds of an LLM's parameters, primarily capture its own knowledge. 
So, we explore the impact of \modelname on the FFN layers to see how \modelname affects these layers in activating knowledge stored in the LLM.

\begin{figure*}[!t]
\centering
\includegraphics[width=\linewidth]{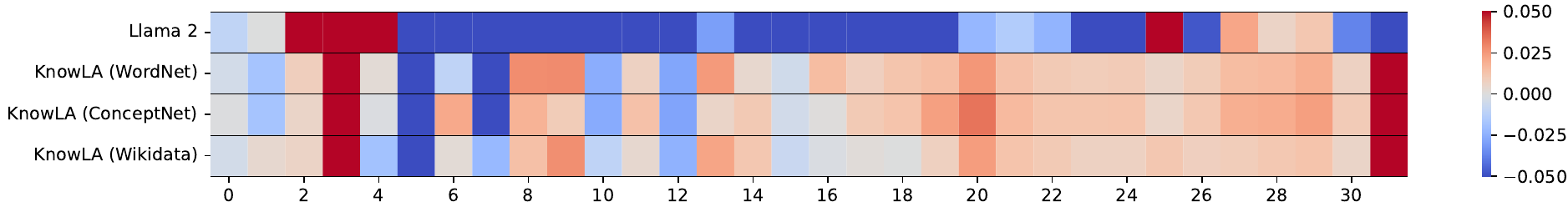}
\caption{
The heatmap indicates the capabilities of \modelname and Llama 2 in capturing knowledge compared to Alpaca2, which is measured by averaging the changes in cosine similarities of the last token representations from 100 queries across all FFN layers.
The x-axis denotes the 32 layers of \llama.}
\label{fig:layer_similarity}
\end{figure*}

We compute the differences between the hidden state representations of the last token before and after each FFN layer in the LLM.
We analyze the trends in differences of all 32 layers after inserting \modelname. 
We use the 100 questions from \triviaqa as queries to explore the knowledge stored in the FFN layers of \llama (7B). 
The last token representation in each input aggregates information from all tokens. 
According to \cite{pmet}, there is a positive correlation between the similarity of hidden states and the consistency of knowledge. 
Intuitively, we believe that higher differences in representations indicate the model's ability to capture more information from the FFN layers. 
Therefore, we extract the representations of the last token before and after each FFN layer and compute the cosine similarities for Llama 2, \modelname, and Alpaca2, which are denoted by $s_{1}$, $s_{2}$, and $s_{3}$, respectively.
Given the token similarities, we further evaluate the capacities of \modelname and Llama 2 in capturing hidden knowledge. 
The capacities are measured by $s_{3}-s_{2}$ and $s_{3}-s_{1}$. 

The results are shown in Figure~\ref{fig:layer_similarity}. 
The red color indicates that the representation of the last token, after introducing \modelname and undergoing the FFN layers, exhibits a greater change compared to that of Alpaca2. 
The blue color shows the opposite.
We think the representations with greater changes capture more internal knowledge. 

After introducing entity embeddings, \modelname enables the LLM to activate richer knowledge at the FFN layers. 
In contrast, \llama captures less knowledge than Alpaca2.
According to the work \cite{ffn1}, lower FFN layers tend to capture shallow knowledge patterns, while higher FFN layers learn more semantic patterns. 
Our \modelname demonstrates enhanced knowledge activation capabilities at the higher layers,
and thus achieves superior results over Alpaca2.
By examining the differences in similarity across the last 16 layers, we find that \modelname (\conceptnet) shows the greatest similarity difference in the three KGs and performs best on \triviaqa. 
This further emphasizes that the introduction of \conceptnet substantially activates more knowledge stored internally in \llama.

\subsection{Impact of KG Embedding Models} \label{impact}
The KG embedding learning models are used to learn entity embeddings \cite{TransE, rescal, rotate, poskhg}. 
We study the impact of embedding learning models for \modelname.
We obtain entity embeddings of \conceptnet by three representative KG embedding models: RESCAL \cite{rescal}, TransE \cite{TransE}, and RotatE \cite{rotate}.
We show the results of \modelname with these embeddings on the \commonsenseqa, \siqa, and \bbh datasets in Table~\ref{tab:kge}.

\begin{table}[!t]
\centering
\resizebox{\linewidth}{!}{\Large
\begin{tabular}{l|cccccc}
\toprule
& \multicolumn{2}{c}{\commonsenseqa} & \multicolumn{2}{c}{\siqa} & \multicolumn{2}{c}{\bbh} \\
\cmidrule(lr){2-3}\cmidrule(lr){4-5}\cmidrule(lr){6-7}
 & Accuracy & Score & Accuracy & Score & Accuracy & Score \\ 
\midrule
RESCAL &\textbf{58.39}	&46.71	&52.10	&44.91	&27.50	&\textbf{25.96} \\
TransE   & \textbf{58.39} & \textbf{48.19}	&\textbf{53.22}	&\textbf{46.81}	&\textbf{30.19} &25.29\\
RotatE & 57.58	&46.05	&52.00	&44.65	&27.31	&24.94\\
\bottomrule
\end{tabular}}
\caption{Comparison of KG embedding learning models on \commonsenseqa, \siqa, and \bbh, which are pre-trained on \conceptnet for \llama.}
\label{tab:kge}
\end{table}

We can observe that the entity embeddings obtained by TransE achieve favorable results. 
This is attributed to the fact that the TransE embeddings have a good generalization ability and are thus more suitable for \llama. 
RotatE employs complex vector representations for entities and obtains subpar results on \llama. 
This suggests that aligning the complex space of entities with the semantic space of \llama during finetuning is challenging, leading to a loss of original entity knowledge.

\subsection{Robustness of \modelname} \label{sect:other}
We evaluate the robustness of \modelname against three factors:
On the foundation model side, we use LLaMA 1 as another LLM. 
On the instruction data side, we finetune \llama using the Vicuna multi-round dialog data \cite{vicuna2023} to get Vicuna2 and KnowLA (Vicuna2). 
On the PEFT method side, we use AdaLoRA \cite{adalora} to replace LoRA and get Alpaca2 (AdaLoRA) and KnowLA (AdaLoRA).
On the rank side, we finetune \llama using the Alpaca data with rank $r=8$ and get Alpaca2 ($r=8$) and \modelname ($r=8$).

\begin{table}[!t]
\centering
\resizebox{.95\columnwidth}{!}{
\begin{tabular}{l|l|cc}
\toprule
& Methods & Accuracy & Score \\ 
\midrule \multirow{2}{*}{LLM side} & Alpaca1 & 56.59 & 46.03 \\
& \modelname (LLaMA 1) & \textbf{57.74} & \textbf{46.81} \\
\midrule \multirow{2}{*}{Data side} & Vicuna2 & 51.52 & 42.31 \\
& \modelname (Vicuna2) & \textbf{53.56} & \textbf{49.09} \\
\midrule \multirow{2}{*}{PEFT side} & Alpaca2 (AdaLoRA) & 57.58 & \textbf{46.67} \\
& \modelname (AdaLoRA) & \textbf{57.66} & 46.30 \\
\midrule \multirow{2}{*}{Rank side} & Alpaca2 ($r=8$) & 56.92 & 46.25 \\
& \modelname ($r=8$) & \textbf{57.74} & \textbf{46.93} \\
\bottomrule
\end{tabular}}
\caption{Results with different LLMs, instruction data, PEFT methods, and ranks on \commonsenseqa}
\label{tab:appendix}
\end{table}

Table~\ref{tab:appendix} lists the performance of the above models on the commonsense reasoning dataset \commonsenseqa. 
We can see that the three \modelname variants still outperform all baselines.
This experiment shows that \modelname is robust and can bring stable improvement when combined with different LLMs, instruction data, PEFT methods, and ranks. 

\section{Conclusion}
In this paper, we propose a knowledgeable adaptation method \modelname. 
It works with LoRA and injects entity embeddings into an LLM in the PEFT process. 
Compared to Alpaca2, which is finetuned with LoRA alone, \modelname with \llama shows better performance on six benchmark datasets.
We show that pre-trained KG embeddings are compatible with \llama. 
Moreover, we find that \modelname can align the KG space and the LLM space, and activate the hidden knowledge related to input in LLMs, thereby achieving improved performance. 

\section*{Limitations}
Currently, our work only incorporates one KG to enhance PEFT. 
As KGs are incomplete by nature, integrating multiple KGs into our method may further improve performance with knowledge fusion and transfer.
Recent work \cite{MultilingualKGCAdaptive} reveals that multi-source KG embeddings are more expressive than the embeddings of a single KG.
We show preliminary results in Appendix~\ref{app:multikg} and will study multi-source \modelname in future work.

We have not attempted other LLMs such as ChatGLM \cite{chatglm} in this work.
In the future, we will consider how to efficiently inject KG knowledge with smaller parameters. 
Additionally, we have observed that, with the introduction of random perturbations, \llama seems to outperform Alpaca2 on some tasks. 
This discovery may provide interesting directions for future research.

\section*{Ethical Considerations}
LLMs may produce incorrect and potentially biased content. 
Experiments show that our method can alleviate this problem to a certain extent, but LLMs will inevitably generate offensive answers. 
Therefore, extreme caution should be exercised if deploying such systems in user-facing applications.
All datasets and models used in this work are publicly available under licenses.

\section*{Acknowledgments}
This work was supported by the National Natural Science Foundation of China (No. 62272219) and the CCF-Tencent Rhino-Bird Open Research Fund.

\bibliography{anthology,custom}

\appendix

\section{Datasets and KGs} \label{app:datasets}
The details of the datasets are described as follows:
\begin{itemize}[itemsep=1pt,topsep=2pt]
    \item In \textbf{\commonsenseqa} \cite{csqa}, each sample consists of a question, five candidate answers, and a correct answer. 
    To run LLMs for \commonsenseqa, we adopt the same setting as in \cite{self-talk} and consider it as a text completion task. We test the LLMs with the validation dataset.

    \item \textbf{\siqa} \cite{siqa} is a QA dataset about social commonsense, where each sample consists of a question, three candidate answers, and a correct answer. 
    To evaluate prompt-based methods, we do not use the provided knowledge in the dataset. 
    The settings are the same as in \commonsenseqa. We test the LLMs with the validation dataset.
        
    \item \textbf{\bbh} \cite{bbh} is a popular benchmark that focuses on tasks challenging for LLMs. 
    To compare scores of different methods on correct answers, we select 15 multiple-choice QA datasets from this benchmark.

    \item \textbf{\webquestionsp} \cite{wqsp} is a KBQA dataset that enhances the WebQuestion dataset by annotating each answer with corresponding SPARQL queries and removing ambiguous, unclear, or unanswerable questions. 
    In this paper, we treat it as a closed-book QA task. 

    \item \textbf{\triviaqa} \cite{triviaqa} includes 95K question-answer pairs authored by trivia enthusiasts, which provide high-quality distant supervision for answering the questions. In this paper, we treat it as a closed-book QA task and select 7,500 questions from \triviaqa to test LLMs.

    \item \textbf{\truthfulqa} \cite{truthfulqa} is a benchmark to measure whether a language model is truthful in generating answers to questions. 
\end{itemize}

The used KGs are introduced as follows:
\begin{itemize}[itemsep=1pt,topsep=2pt]
    \item \textbf{\wordnet} \cite{WordNet} is a lexical KG in English. Nouns, verbs, adjectives, and adverbs are arranged into synsets, each denoting a separate notion.
 
    \item \textbf{\conceptnet} \cite{ConceptNet} is a multi-lingual conceptual KG of things people know and computers should know.

    \item \textbf{\wikidata} \cite{wikidata} is a factual KG across diverse domains. 
    It encompasses various entity types, including individuals, places, concepts, etc. 
\end{itemize}

\section{Knowledge-Unrelated Tasks} \label{app:side_effect}
We analyze the side effects of \modelname on knowledge-unrelated tasks. 
In this experiment, five knowledge-unrelated tasks from BBH are picked.
The results in Table \ref{tab:side_effect} show that even if these tasks are knowledge-unrelated, our \modelname can still improve the LLM. 
This is due to the enhanced ability of the LLM to activate its own knowledge.

\begin{table}
\centering
\resizebox{0.99\columnwidth}{!}{
\begin{tabular}{l|cc}
\toprule
Datasets & \alpaca ($r=16$) & \modelname (\conceptnet) \\ 
\midrule
Temporal sequences & 14.80 & 15.20\\
Date understanding & 72.00 & 73.20\\
Geometric shapes & \ \ 9.20 & 19.20\\
Snarks & 51.12 & 53.37 \\
Logical deduction & 35.20 & 36.40\\
\bottomrule
\end{tabular}}
\caption{Results on knowledge-unrelated tasks}
\label{tab:side_effect}
\end{table}

\section{Additional Latency on Efficiency} \label{app:efficiency}
Retrieving the embeddings of related entities during each finetuning step would slow down the training process. 
We move it to the data processing step. 
We use eight workers to process 50,538 training samples in parallel. 
During inference, we compare the overall inference time of \modelname (\conceptnet) and Alpaca2 on \commonsenseqa using an A6000 GPU card. 
Table \ref{tab:time} shows the results.

\begin{table}
\centering
\resizebox{0.99\columnwidth}{!}{
\begin{tabular}{l|cc}
\toprule
Models & Data processing & Inference \\ 
\midrule
Alpaca2 & \ \ 9 s & 19.02 min.\\
\modelname (\conceptnet) & 16 s & 19.20 min.\\
\bottomrule
\end{tabular}}
\caption{Time overhead of Alpaca2 and \modelname}
\label{tab:time}
\end{table}

Alpaca2 spends 9 seconds on data processing, while \modelname (\conceptnet) spends 16 seconds. 
During inference, \modelname (\conceptnet) takes 19 minutes and 12 seconds, while Alpaca2 takes 19 minutes and 1 second. 
We believe that the additional latency caused by \modelname is tolerable compared to the performance boost.

\section{Robustness to Different Prompts}
We try different prompts to evaluate the robustness of \modelname. 
Table~\ref{tab:prompt} compares the accuracy of Alpaca2 ($r=16$) and \modelname on \commonsenseqa with different prompts. 
\modelname outperforms Alpaca2 on all prompts, indicating its good robustness.
We use the first prompt in the main experiments due to its superior performance. 

\begin{table}
\centering
\resizebox{\columnwidth}{!}{
\begin{tabular}{p{6.1cm}|cc}
\toprule
\multirow{2}{*}{Prompts} & Alpaca2 & \modelname \\ 
& ($r=16$) & (\conceptnet) \\
\midrule
Below is an instruction that describes a task, paired with an input that provides further context. Choose a correct answer that appears in the candidate answers. & 56.92 & 58.39 \\
\midrule
Below is an instruction that describes a task, paired with an input that provides further context. Please answer the following question. & 52.74 & 54.95 \\
\midrule
Below is an instruction that describes a task, paired with an input that provides further context. Give an answer that appropriately completes the question. & 53.73 & 56.10 \\
\midrule
Please answer the following question. & 55.20 & 56.35 \\
\bottomrule
\end{tabular}}
\caption{Accuracy of \modelname when using different prompts on \commonsenseqa}
\label{tab:prompt}
\end{table}

\section{Discussion on Extension of \modelname} \label{app:disscussion}
We hereby discuss the extension of \modelname to integrate multiple KGs for PEFT and incrementally incorporate knowledge updates in a KG.

\subsection{Multiple KGs}\label{app:multikg}
To leverage multiple KGs and try to benefit from their potential knowledge transfer,
we design a simple baseline that merges different KGs into a large graph using entity alignment \cite{BootEA} and learns entity embeddings from this large KG. 
If there is no entity alignment, we use entity embeddings from these KGs simultaneously. 
For balanced training, we limit the maximum number of related entities from each KG to two, and each token has up to six entity embeddings. 

\begin{table}
\centering
{\small
\begin{tabular}{l|cc}
\toprule
Methods & Accuracy & Score \\ 
\midrule
\modelname (\wordnet) & 58.07 & \textbf{48.35} \\
\modelname (\conceptnet) & \textbf{58.39} & 48.19 \\
\modelname (\wikidata) & 57.90 & 47.39 \\
\modelname (multiple KGs) & 57.61 & 47.24 \\
\bottomrule
\end{tabular}}
\caption{Results of multiple KGs on \commonsenseqa}
\label{tab:multikg}
\end{table}

In this experiment, we merge \wordnet, \conceptnet, and \wikidata. 
According to Table \ref{tab:multikg}, the accuracy of this baseline on \commonsenseqa is 57.61, which is slightly lower than the result of \modelname (\conceptnet). 
We think this straightforward baseline may not effectively leverage knowledge transfer between KGs. 
A more promising mechanism, e.g., Mixture of Experts \cite {shen2023mixtureofexperts}, is necessary to combine multiple KGs. 
\modelname is adaptable to integrate this improvement. 
We leave this direction for future work.

\subsection{Knowledge Updates}
Knowledge updates for \modelname require the support of incremental learning for KGs and LLMs. 
When some new entities and triples are added to a KG, only a small number of parameters need to be re-trained to complete knowledge updates by using lifelong KG embedding learning \cite{lifelong} and continual PEFT \cite{O-LoRA}.

\end{document}